\documentclass{article}
\usepackage{spconf,amsmath,graphicx}
\usepackage{color}
\usepackage{multirow}
\usepackage{amsmath,amssymb}
\usepackage{graphicx}
\usepackage {booktabs}
\usepackage[numbers]{natbib}

\title{Cascaded Interaction with Eroded Deep Supervision for\\ Salient Object Detection}
\name{Hewen Xiao$^{1}$, Jie Mei$^{2}$, Guangfu Ma$^{2}$, Weiren Wu$^{1}$ 
}
\address{$^{1}$Institute of Space Science and
Applied Technology,
Harbin Institute  of  Technology, Shenzhen, China \\$^{2}$School of Mechanical Engineering and Automation,
 Harbin Institute  of  Technology, Shenzhen, China}

\begin{document}
%
\maketitle
\begin{abstract}
Deep convolutional neural networks have been widely applied in salient object detection and have achieved remarkable results in this field.  However, existing models suffer from information distortion caused by interpolation during up-sampling and down-sampling. In response to this drawback, this article starts from two directions in the network: feature and label. On the one hand, a novel cascaded interaction network with a guidance module named global-local aligned attention (GAA) is designed to reduce the negative impact of interpolation on the feature side. On the other hand, a deep supervision strategy based on edge erosion is proposed to reduce the negative guidance of label interpolation on lateral output. Extensive experiments on five popular datasets demonstrate the superiority of our method. 
\end{abstract}
\section{Introduction}
\label{sec:intro}

Salient object detection (SOD) aims to identify the most visually prominent objects in an image. 
Traditional SOD approaches usually utilize hand-crafted cues, such as smoothing techniques~\cite{Eun2015SOD,pixel2021xu} and global cues \cite{cheng2015global}. These features may not describe the complex image scene or can not adapt to the new scene or object. These facts lead to poor generalization ability, and therefore traditional SOD methods fall into a bottleneck.
Recently, Convolutional Neural Networks (CNN) have been developed for salient object detection and show their powerful feature extraction ability. 
Among them, early work \cite{wang2019iterative,wu2019cascaded} mostly adopted an iterative or stage-wise manner. Some later methods \cite{liu2021rethinking,liu2019simple,liu2022poolnet+} focused on designing multi-scale feature-extracting modules. Attention mechanisms are introduced to enhance the network representation \cite{wang2019salient,liu2021rethinking}.

Although CNN-based methods achieve significant results compared to traditional methods, there are still some challenges in making more accurate detection. 
To improve the network representation ability, a very important means is to increase the depth of the neural network to learn as many hierarchical features as possible. However, the feature information will be lost with the increase in network depth. Hence most existing deep models for SOD adopted the encoder-decoder paradigm~\cite{qin2019basnet,liu2019simple}, and connected the encoder directly to the corresponding level in the decoder to ensure the transfer of feature maps and information. Since the scale of the salient object to be detected is often different, the receptive field size required is also different. Therefore, multi-scale feature fusion is further proposed for the network to obtain features in different scales, which can be simply divided into feature and label levels. However, in either case,  the fusion between the different levels inevitably involves the use of spatial interpolation. When the image is up-sampled and then down-sampled, the essential information of the image is lost seriously. If we reverse the sampling order, the situation will be more serious. In this paper, we focus on the feature level and label level to address this challenge.

\begin{figure}[t]
    \centering
    \includegraphics[width=0.48\textwidth]{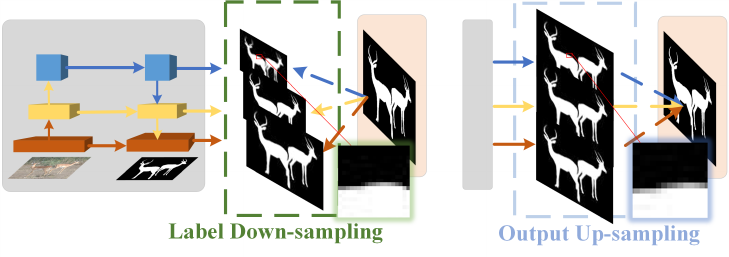}
\caption{The example of interpolation operations in deep supervision. The loss of information caused by interpolation can not be avoided in either Down-sampling or Up-sampling. }
\label{fig1}
\end{figure}

\begin{figure*}[htbp]
    \centering
    \includegraphics[width=0.9\textwidth]{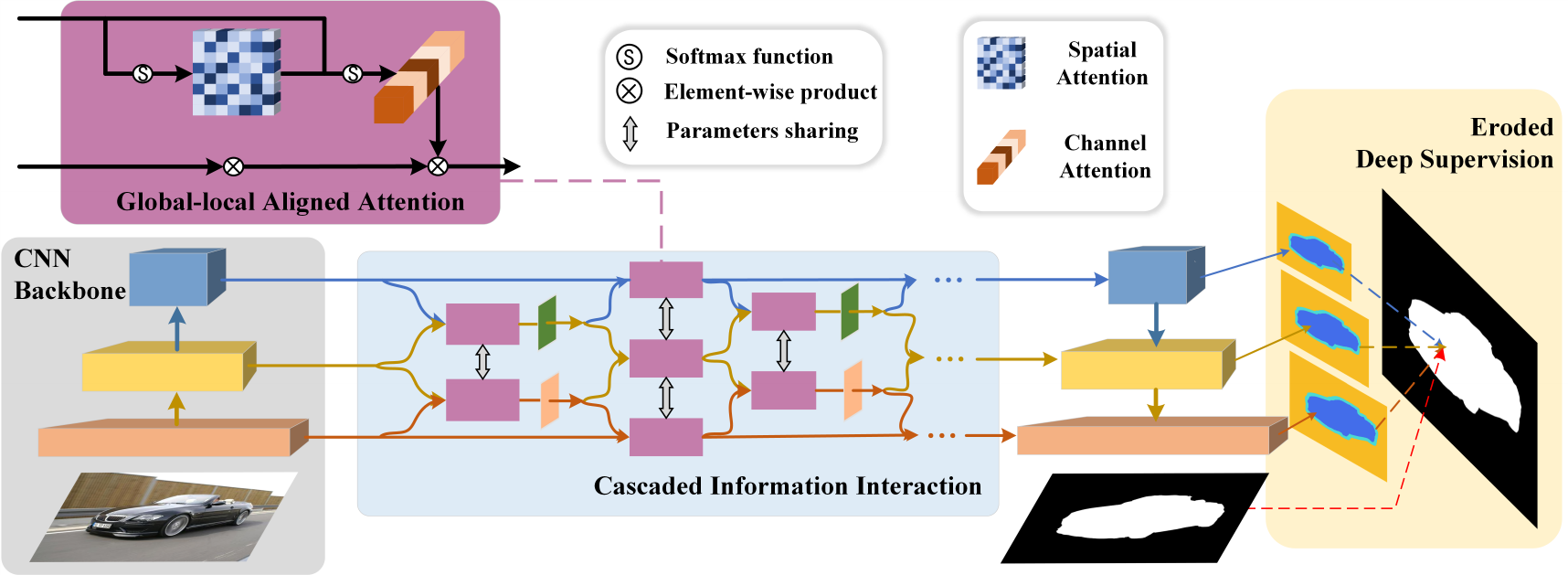}
    \caption{An overview of the proposed cascaded interaction network framework.
We use cascaded information interaction and eroded deep supervision to deal with the errors introduced by interpolation in interaction features and supervised labels, respectively. Among them, the GAA module is adopted as the information interaction, as shown in the top-left corner.} 
    \label{fig2}
\end{figure*}

For information loss caused by interpolation operations at the feature level, combining multi-scale information exchange at the filter level can better help mitigate this negative impact than the feature level~\cite{liu2021rethinking}. However, the existing multi-scale information interaction is still a single-level structure. We propose a new cascaded interaction network, for deepening the network to continuously refine the extraction of knowledge and making interaction a richer feature, with a higher dimension of information interaction to deal with challenging tasks. 
Unlike the previous cascaded methods for SOD~\cite{wu2019cascaded}, our cascading goal is to focus on multi-scale information exchange between encoders and decoders. Moreover, we design a new GAA module that shares parameters at different scales at each level to achieve information interaction at the filter level. This design not only reduces the number of parameters but also encourages the shared filters to adapt to inputs at different scales, to avoid the negative impact of spatial interpolation.

Moreover, deep supervision is used by adding side-output layers to increase direct supervision of the earlier layers, rather than the final output~\cite{hou2016deeply}. Deep supervision can not only alleviate the problem of gradient vanishing progressively through some “auxiliary” objective functions, but also guide the network to better fit the objective gradually. However, existing results using deep supervision are also affected by the interpolation problem. As shown in Fig. \ref{fig1}, whether the label is sampled above or below the lateral output, the edge value of the label will be inaccurate due to the interpolation. To overcome this challenge, we design a deep supervision strategy based on edge erosion.

In summary, our main contributions include: (a) a novel cascaded network for multi-scale information interactions; (b) a new GAA attention module that encourages filters to learn multi-scale feature expression; (c) an eroded deep supervision strategy to reduce the error caused by inaccurate label edge and (d) experimental results on five popular datasets which demonstrate the effectiveness of the proposed method.

\section{Proposed Method}
\label{sec:Method}
\subsection{Cascaded Interaction Network}
Our proposed method is based on the widely used encoder-decoder structure, which uses a bottom-up branch to extract multi-scale features and a top-down one to combine these features. Among them, the low-level features contain more local textures and patterns, while the deeper-level features reflect the position of the whole object. For extracting each scale feature based on different backbones, we add one convolutional layer with a kernel size of 1 to unify the number of channels. Therefore, the features extracted from the encoder with unified channels could be described as $\mathcal{E} = \{E_i, 1\le i \le I\}$ (note that $I$ is usually selected as 5).

Considering the loss of information caused by feature interpolation, we introduce cascaded multi-scale information interaction at the filter level. More importantly, a generally accepted perception is that the network performs better when going deeper. This phenomenon also applies to the interaction between features. Then in this paper, we insightfully extend the number of levels of interaction. As shown in Fig. \ref{fig2}, between the bottom-up and top-down pathways, the light blue rectangle denotes the cascaded information interaction, and the resulting information is passed on to subsequent one for deeper fusion. Given the channel unified feature maps from the encoder $\mathcal{E}$, the features delivered to the decoder $\mathcal{D} = \{D_j, 1\le j \le J\}$ can be got by cascaded interactors as
\begin{equation}
D_{j}=\mathbb{I}^{q}\left(E_{k}, \ldots, E_{m}\right), \quad 1 \leq j \leq J, \quad 1\leq k \leq m \leq I
\end{equation}
where $\mathbb{I}$ denotes the feature fusion in each interaction level, 
and $q$ indicates the number of function actions, here specifically meaning the number of cascading interaction levels.

\subsection{Global-local Aligned Attention}
Existing modules are usually designed for single-scale input.  While the above proposed cascaded network obviously needs to process multi-input for multi-scale information interaction. Therefore, we further introduce the GAA module, which tells the network where to focus and also enhances the feature expression of key areas. As shown in the pink area in Fig. \ref{fig2}, the spatial attention $M_S$ and channel attention $M_C$ mechanisms \cite{woo2018cbam} are sequentially used to obtain the refined feature. The network has the ability to learn "What" and "Where", allowing it to better understand which information needs to be emphasized and which information needs to be suppressed, for better guiding the global information to the relative local one.
Specifically, the softmax operation is added before each attention guide.


Note that in every single level interaction in our CINet, we use a solo GAA, that is sharing parameters in every interaction stage. Its advantage is that it can not only save parameters but also strengthen the filter-level interaction between different scale information at each level.
 
\subsection{Eroded Deep Supervision Loss}
Binary cross entropy (BCE) loss is one of the most widely used loss
because of its robust performance in binary classification and segmentation tasks, which
accumulates the loss for each pixel in images independently, defined as
\begin{equation}
	L_{bce}(x,y)=-\frac{1}{n} \sum_{k=1}^n [y_klog(x_k)+(1-y_k)log(1-x_k)], 
\end{equation}
where $x$ and $y$ denote the predicted map 
and the ground truth, respectively, 
while $k$ is the index of pixels and $n$ is the number of pixels in $x$.
However, BCE loss lacks an estimate of the overall image. To compensate for this, intersection over union (IoU) loss is introduced to consider
the similarity of the image entirely, 
which is given by
\begin{equation}
	L_{iou}(x,y) = 1-\frac{\sum_{k=1}^n (y_k\cdot x_k)}{\sum_{k=1}^n (y_k+x_k-y_k\cdot x_k)}.
\end{equation}

\begin{figure}[t]
\begin{minipage}[b]{.48\linewidth}
  \centering
  \centerline{\includegraphics[width=4.cm]{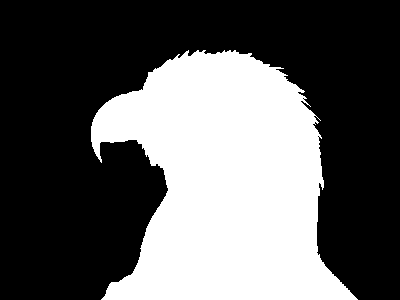}}
  \centerline{(a) Original Label}\medskip
  \label{fig3a}
\end{minipage}
\hfill
\begin{minipage}[b]{.48\linewidth}
  \centering
  \centerline{\includegraphics[width=4.cm]{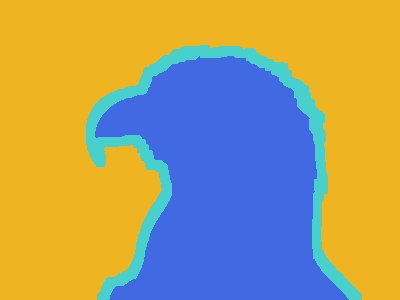}}
  \centerline{(b) Eroded Label}\medskip
  \label{fig3b}
\end{minipage}
\caption{(a) represents $\mathcal{X}$; The yellow and blue parts in (b) represent $\mathcal{C}$, while the light blue part represents $\mathcal{E}$. }
\label{fig3}
\end{figure}

\begin{table}
\setlength\tabcolsep{4mm}
\centering
\caption{Ablation study on  cascaded levels.}
\begin{tabular}{ccccc}
\hline
 & 0 &1 & 2 & 3 \\
\hline
$F_\beta$$\uparrow$ & 0.848 & 0.878 &  \textbf{0.888} & 0.885\\
\hline
MAE$\downarrow$ & 0.052 &  0.042 &  \textbf{0.038} & 0.039 \\
\hline
\end{tabular}
\label{t1}
\end{table}

\begin{table}[t]
\setlength\tabcolsep{5mm}
\centering
\caption{Ablation study on global-local aligned attention.}
\begin{tabular}{ccccc}
\hline
 & Spatial & Channel & GAA \\
\hline
$F_\beta$$\uparrow$ & 0.879 & 0.871 &  \textbf{0.888} \\
\hline
MAE$\downarrow$ & 0.041 &  0.046 &  \textbf{0.038} \\
\hline
\end{tabular}
\label{t2}
\end{table}

\begin{table}[t]
\setlength\tabcolsep{5mm}
\setlength{\abovecaptionskip}{0pt}%
\setlength{\belowcaptionskip}{10pt}%
\centering  
\caption{Ablation study on deep supervision type.}
\begin{tabular}{ccccc}
\hline
 & w/o & Normal & Eroded \\
\hline
$F_\beta$$\uparrow$ & 0.882 & 0.859 &  \textbf{0.888} \\
\hline
MAE$\downarrow$ & 0.039 &  0.047 &  \textbf{0.038} \\
\hline
\end{tabular}
\label{t3}
\end{table}

\begin{table*}
\label{all_comp}
\small
\setlength\tabcolsep{3mm}
\begin{center}
\begin{tabular}{|l|cc|cc|cc|cc|cc|}
\hline
\multirow{2}*{Method}  &
        \multicolumn{2}{c|}{ECSSD } & \multicolumn{2}{c|}{PASCAL-S } & \multicolumn{2}{c|}{DUT-OMRON} & \multicolumn{2}{c|}{HKU-IS} & \multicolumn{2}{c|}{DUTS-TE}
        \\ \cline{2-11} 
    & $F_\beta$~$\uparrow$ & MAE~$\downarrow$  & $F_\beta$~$\uparrow$ & MAE~$\downarrow$  & $F_\beta$~$\uparrow$ & MAE~$\downarrow$  & $F_\beta$~$\uparrow$ & MAE~$\downarrow$ & $F_\beta$~$\uparrow$ & MAE~$\downarrow$\\
\hline
        $\text{MLMS}$~\cite{wu2019mutual} & 0.930 & 0.045 & 0.853 & 0.074 & 0.793 & 0.063 & 0.922 & 0.039 & 0.854 & 0.048 \\  
        $\text{PAGE}$~\cite{wang2019salient} & 0.931 & 0.042 & 0.848 & 0.076 & 0.791 & 0.062 & 0.920 & 0.036 & 0.838 & 0.051 \\ 
        $\text{CapSal}$~\cite{zhang2019capsal} & - & - & 0.862 & 0.073 & - & - & 0.889 & 0.058 & 0.844 & 0.060 \\ 
        $\text{ICTB}$~\cite{wang2019iterative} & 0.938 & 0.041 & 0.855 & 0.071 & 0.811 & 0.060 & 0.925 & 0.037 & 0.855 & 0.043\\
        $\text{CPD}$~\cite{wu2019cascaded}  & 0.939 & 0.037 & 0.859 & 0.071 & 0.796 & 0.056 & 0.925 & 0.034 & 0.865 & 0.043 \\ 
        $\text{BASNet}$~\cite{qin2019basnet} & 0.942 & 0.037 & 0.857 & 0.076 & 0.811 & 0.057 & 0.930 & 0.033 & 0.860 & 0.047 \\ 
        $\text{PoolNet}$~\cite{liu2019simple} & 0.944 & 0.039 & 0.865 & 0.075 & 0.830 & 0.055 & 0.934 & 0.032 & 0.886 & 0.040 \\
        $\text{CSNet}$~\cite{gao2020sod100k}  & 0.944 & 0.038 & 0.866 & 0.073 & 0.821 & 0.055 & 0.930 & 0.033 & 0.881 & 0.040 \\
        $\text{ITSD}$~\cite{zhou2020interactive}  & 0.947 & \textbf{0.035} & \textbf{0.871} & 0.066 & 0.823 & 0.061 & 0.933 & 0.031 & 0.883 & 0.041 \\
        \hline
        $\textbf{CINet(Ours)}$ & \textbf{0.948} & 0.036 & \textbf{0.871} & \textbf{0.064} & \textbf{0.834} & \textbf{0.053} & \textbf{0.939} & \textbf{0.030} & \textbf{0.888} & \textbf{0.038}\\
\hline
\end{tabular}
\end{center}
\caption{Comparisons of our method with other state-of-the-art methods on five popular SOD benchmarks.}
\label{t4}
\end{table*}

However, due to the impact of interpolation on deep supervision methods as shown in Fig. \ref{fig1}, the edge values of labels may be inaccurate due to interpolation. To solve this problem, we redesign the loss function based on edge erosion in the side output layer of deep supervision. Here we define $\mathcal{C} = \mathcal{X}-\mathcal{E}$, where $\mathcal{X}$ (as shown in Fig. \ref{fig3}(a)) is the set of all pixels of the label, and $\mathcal{E}$ (the light blue part of Fig. \ref{fig3}(b)) is the set of pixels in the eroding edge part of the label, and $\mathcal{C}$ (the yellow and blue parts in Fig. \ref{fig3}(b)) includes all the remaining pixels after the edge erosion is eliminated from the label. While the new losses are computed only in the remaining set $\mathcal{C}$. Here we take the example of BCE loss at the kth pixel point in the set $\mathcal{C}$, which can be rewritten as:
\begin{equation}
	l_{bce}^{E}(x_k,y_k)=-[y_klog(x_k)+(1-y_k)log(1-x_k)],
\end{equation}
where $x_k,y_k \in \mathcal{C}$.
We generalize the loss of a single pixel to the set level, and use $l_{bce}^{E}$ and $l_{iou}^{E}$ to represent the BCE loss and IoU loss of set $\mathbb{C}$, respectively. The training loss is then defined as the sum of all outputs:
\begin{equation}
Loss=L_{bce}(x,y)+L_{iou}(x,y)+\sum_{m=1}^M \alpha_{m}\cdot l^{E_m}_{bce}+\sum_{m=1}^M \beta_{m}\cdot l^{E_m}_{iou}
\end{equation}
where $l^{E_m}$ is the loss of the m-th side output, $M$ denotes the total number of the outputs, $\alpha$ and $\beta$ represent the weights of eroded BCE loss and IoU loss for deep supervision, and $l_{bce}(x,y),l_{iou}(x,y)$ are the final output loss. Since it is not affected by interpolation, we keep the original form here.

\section{Experimental Results}
\subsection{Experiment Setup}
\textbf{Datasets and Evaluation Metrics.\quad} 
The datasets we used for evaluation include 
five popular datasets: ECSSD \cite{yan2013hierarchical}, 
PASCAL-S \cite{li2014secrets}, 
DUT-OMRON \cite{yang2013saliency}, 
HKU-IS \cite{li2015visual}  
and DUTS-TE \cite{wang2017learning}.
For training the model, 
we adopt DUTS-TR \cite{wang2017learning} dataset 
in all the experiments following common SOD works.
In addition, we use two widely used metrics: 
F-measure score ($F_\beta$) and  
mean absolute error ($MAE$).
Note that ablation studies are conducted on DUTS-TE~\cite{wang2017learning} dataset.

\noindent \textbf{Implementation Details.\quad}
%
%
The backbone we utilize is the ResNet-50 \cite{He2016} pre-trained on 
the ImageNet dataset \cite{krizhevsky2012imagenet}.
The input images are resized to $352\times352$ in both training and testing phrases.
%
%
For the optimization, we select the stochastic gradient descent (SGD) optimizer 
with a learning rate of 0.005, a momentum of 0.9, and a weight decay of 5e-5.
%
%
The training process lasts 32 epochs 
with a batch size of 30.

%
%
%

\subsection{Ablation Studies}


\noindent \textbf{Effectiveness of Cascaded Interaction Network.\quad}As shown in the first column of Table \ref{t1}, the number "0" refers to the network without any interaction,  whose feature transfer is only based on short connections. While other numbers represent the cascaded number for feature interaction. 
After using one layer of feature interaction, $F_\beta$ increased by 0.03 and MAE decreased by 0.01. Feature interaction reduced information loss caused by interpolation and improved network performance. Further increasing the number of interaction layers, $F_\beta$ increased to 0.888, and MAE decreased to 0.038, indicating that the two-layer interaction improved the richness of features and further improved detection accuracy. Observing the interaction of three-layer features, we found that blindly increasing the number of cascading layers does not significantly improve network performance but may actually decrease it. Therefore, to balance both network performance and parameter complexity, our other ablation experiments are based on the two-level cascaded interaction by default unless special explanations. 

\noindent \textbf{Effectiveness of Global-local Aligned Attention.\quad} As shown in Table \ref{t2}, we compared the experimental results in the presence or absence of spatial and channel attention. 
The channel attention focuses on the importance and relevance of key information, while the spatial attention focuses on which locations have key information. Observing from the GAA experiment, $F_\beta$ reached 0.888 and MAE decreased to 0.038, both of which were superior to those used alone, proving the effectiveness of the GAA module.

\noindent \textbf{Effectiveness of Eroded Deep Supervision.\quad} We conducted the following experiments separately: without deep supervision (w/o), with normal deep supervision (Normal), and with eroded deep supervision (Eroded). As shown in Table \ref{t3}, we found that the results of Normal deep supervision were inferior to w/o, as interpolation resulted in inaccurate label edge values, leading to supervisory errors in the lateral output. Eroded deep supervision eliminates edge errors during side output and improves detection accuracy.

\subsection{Comparisons to the State-of-the-Arts}
\begin{figure}[htbp]
\centering
 \begin{minipage}{0.13\linewidth}
 	\vspace{1pt}
 	\centerline{\includegraphics[width=\textwidth]{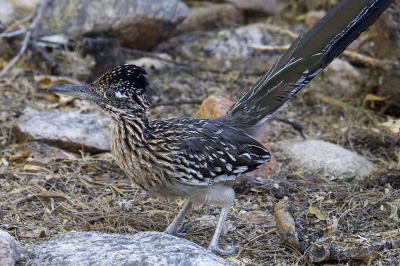}}
 	\vspace{1pt}
 	\centerline{\includegraphics[width=\textwidth]{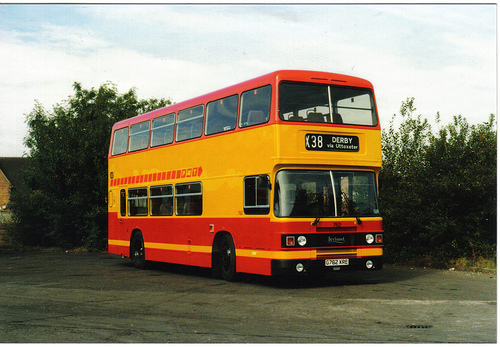}}
 	\vspace{1pt}
 	\centerline{\includegraphics[width=\textwidth]{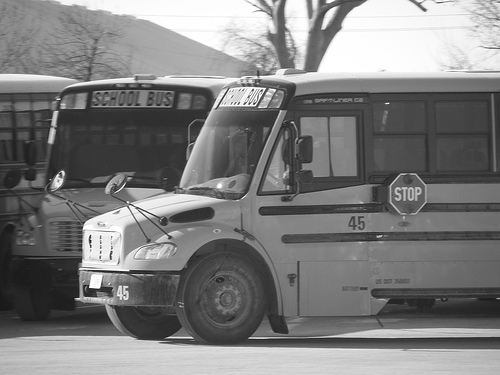}}
 	\vspace{1pt}
 	\centerline{\includegraphics[width=\textwidth]{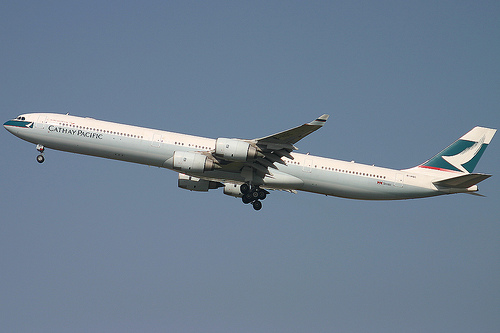}}
 	\vspace{1pt}
 \end{minipage}
 \begin{minipage}{0.13\linewidth}
	\vspace{1pt}
	\centerline{\includegraphics[width=\textwidth]{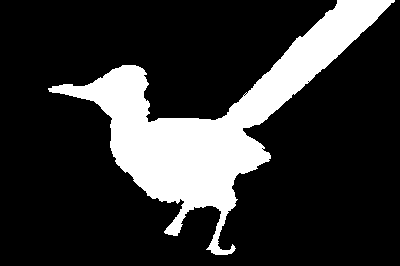}}
	\vspace{1pt}
 	\centerline{\includegraphics[width=\textwidth]{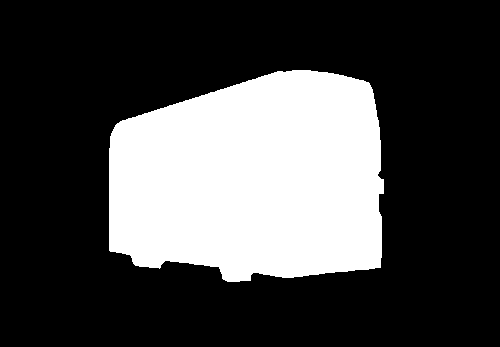}}
	\vspace{1pt}
	\centerline{\includegraphics[width=\textwidth]{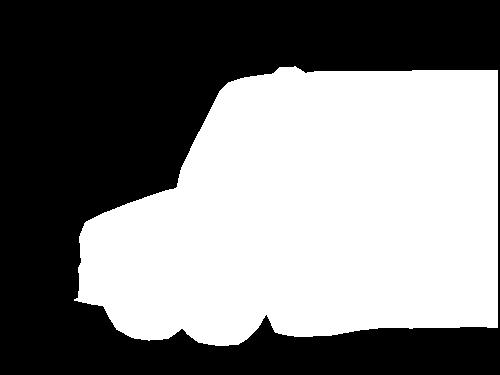}}
	\vspace{1pt}
	\centerline{\includegraphics[width=\textwidth]{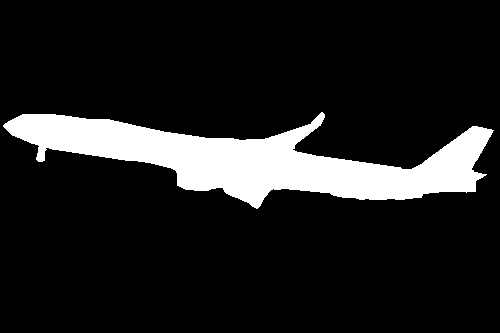}}
	\vspace{1pt}
\end{minipage}
\begin{minipage}{0.13\linewidth}
	\vspace{1pt}
	\centerline{\includegraphics[width=\textwidth]{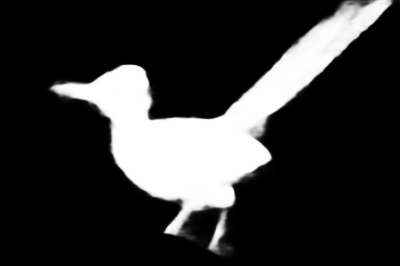}}
	\vspace{1pt}
	\centerline{\includegraphics[width=\textwidth]{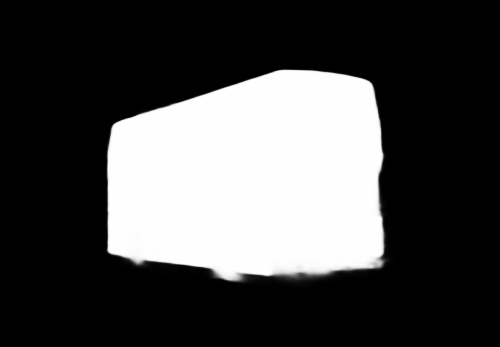}}
	\vspace{1pt}
	\centerline{\includegraphics[width=\textwidth]{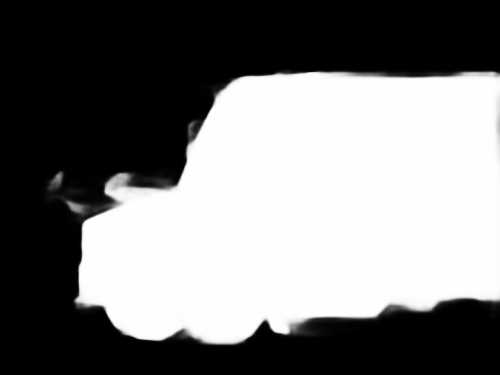}}
	\vspace{1pt}
	\centerline{\includegraphics[width=\textwidth]{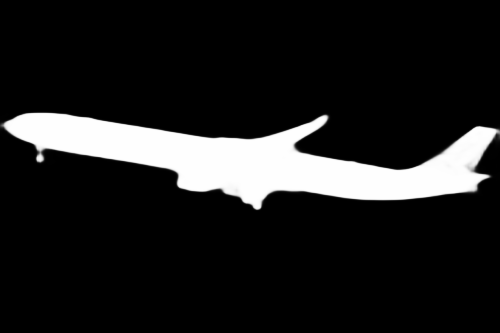}}
	\vspace{1pt}
\end{minipage}
 \begin{minipage}{0.13\linewidth}
 	\vspace{1pt}
 	\centerline{\includegraphics[width=\textwidth]{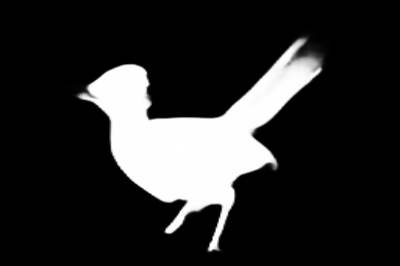}}
 	\vspace{1pt}
 	\centerline{\includegraphics[width=\textwidth]{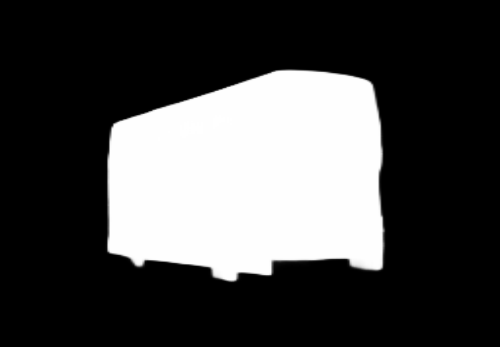}}
 	\vspace{1pt}
 	\centerline{\includegraphics[width=\textwidth]{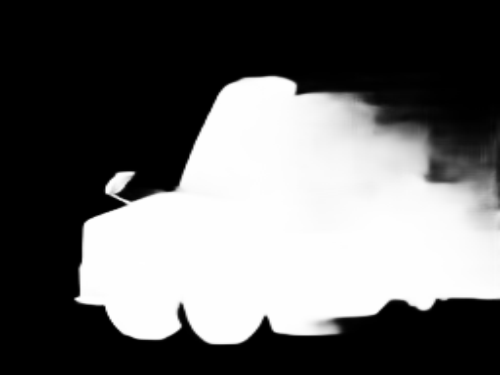}}
 	\vspace{1pt}
 	\centerline{\includegraphics[width=\textwidth]{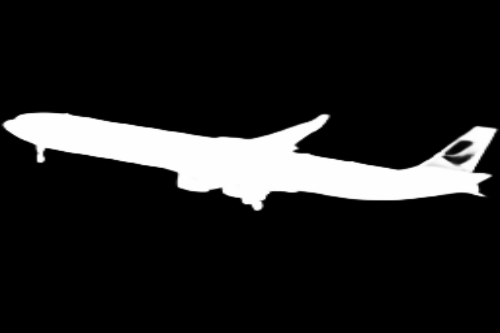}}
 	\vspace{1pt}
 \end{minipage}
 \begin{minipage}{0.13\linewidth}
	\vspace{1pt}
	\centerline{\includegraphics[width=\textwidth]{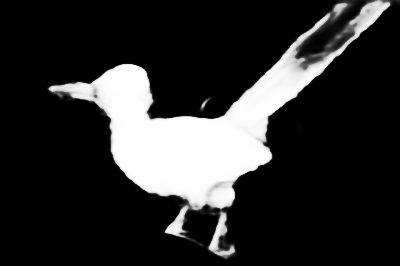}}
	\vspace{1pt}
 	\centerline{\includegraphics[width=\textwidth]{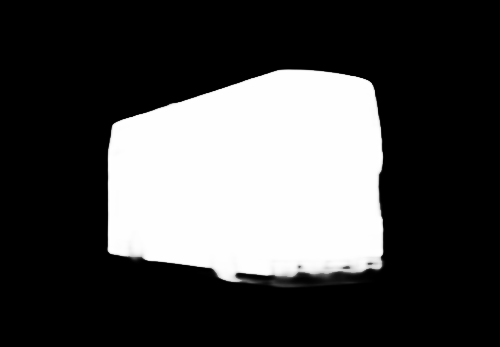}}
	\vspace{1pt}
	\centerline{\includegraphics[width=\textwidth]{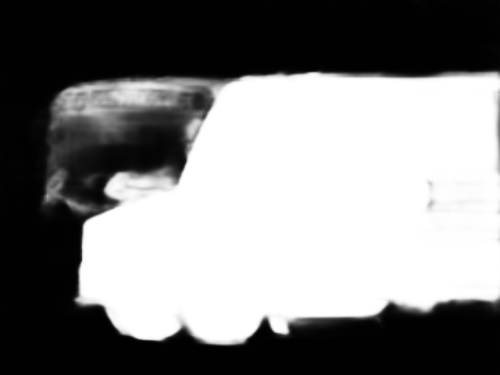}}
	\vspace{1pt}
	\centerline{\includegraphics[width=\textwidth]{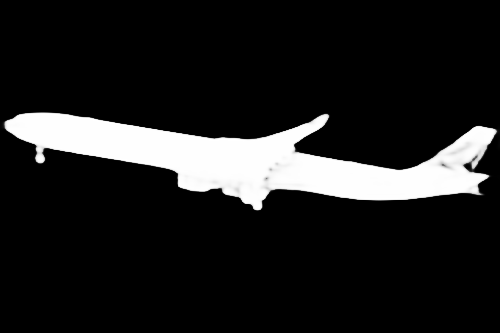}}
	\vspace{1pt}
\end{minipage}
\begin{minipage}{0.13\linewidth}
	\vspace{1pt}
	\centerline{\includegraphics[width=\textwidth]{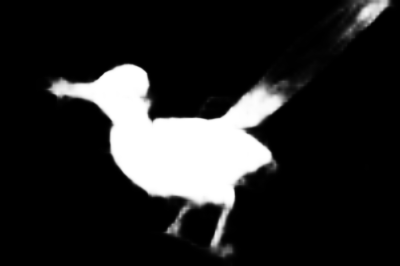}}
	\vspace{1pt}
	\centerline{\includegraphics[width=\textwidth]{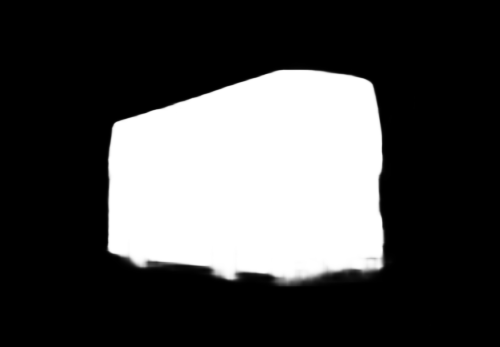}}
	\vspace{1pt}
	\centerline{\includegraphics[width=\textwidth]{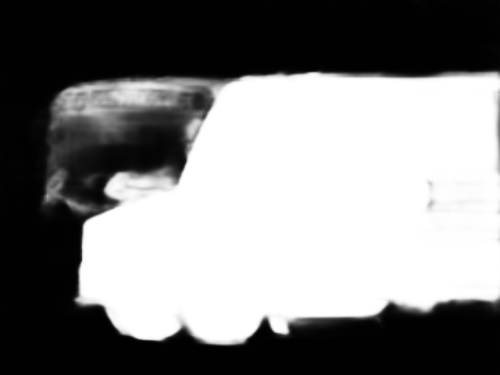}}
	\vspace{1pt}
	\centerline{\includegraphics[width=\textwidth]{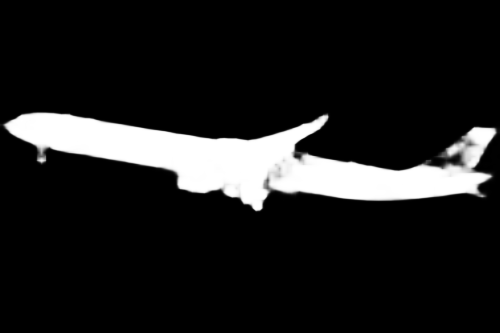}}
	\vspace{1pt}
\end{minipage}
\begin{minipage}{0.13\linewidth}
	\vspace{1pt}
	\centerline{\includegraphics[width=\textwidth]{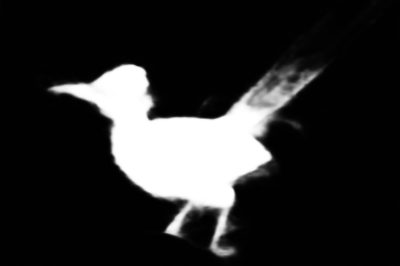}}
	\vspace{1pt}
	\centerline{\includegraphics[width=\textwidth]{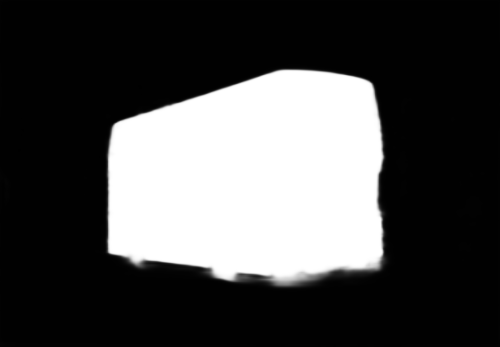}}
	\vspace{1pt}
	\centerline{\includegraphics[width=\textwidth]{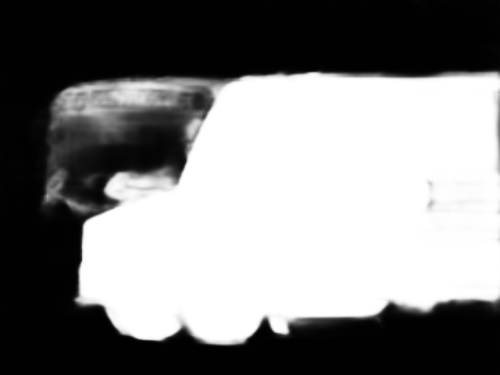}}
	\vspace{1pt}
	\centerline{\includegraphics[width=\textwidth]{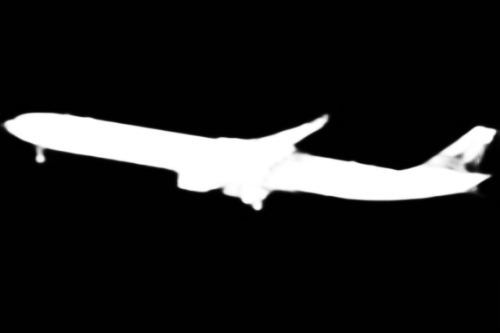}}
	\vspace{1pt}
\end{minipage}
\caption{Visual comparison of saliency maps with state-of-the-art methods. From left to right: Input image, Ground truth, CINet(Ours), BASNet, ITSD , CSNet, and PoolNet. 
}
\label{fig5}
\end{figure}

We compared the proposed CINet with 9 state-of-the-art salient object detection methods.
For a fair comparison, all methods are based on the ResNet-50 backbone. We quantitatively compare the obtained results as shown in Table \ref{t4}. It can be seen that our method performs well compared to other methods corresponding to all datasets on $F_\beta$ and MAE.
We also give some saliency maps of our method in Fig. \ref{fig5}. From this, we can see that our method can produce more accurate results with clear boundaries and uniform highlights.

\section{Conclusion}
In this paper, we have proposed a novel salient object detection
method named cascaded interaction network. 
It introduces multi-scale feature interaction and combines cascading networks to enhance the richness of feature interaction and minimize information loss during the bottom-up process. In addition, we propose a GAA module to allow the network to better fulfill information interaction and introduce an edge erosion strategy to reduce the impact of interpolation on edge information. 
The experimental results on five datasets show that our proposed method outperforms the most advanced methods under different evaluation indicators.

\bibliographystyle{unsrtnat}
\bibliography{strings,refs}

\end{document}